\documentclass[letterpaper, 10 pt, conference]{ieeeconf}  

\IEEEoverridecommandlockouts                              

\usepackage{graphicx}
\usepackage{amssymb,amsmath,physics}
\usepackage{array,booktabs,arydshln}
\usepackage[cal=boondox,scr=boondoxo]{mathalfa}
\usepackage{algorithm}
\usepackage{algorithmic}
\usepackage[utf8]{inputenc}
\usepackage[english]{babel}
\usepackage{xcolor}
\definecolor{green}{RGB}{3,112,15}
\definecolor{yellow}{RGB}{255,140,0}
\usepackage{hyperref}
\usepackage{cite}
\usepackage{subfig}

\title{\LARGE \bf
Extended Abstract: \\Motion Planners Learned from Geometric Hallucination 
}

\author{Xuesu Xiao$^{\dagger}$, Bo Liu$^{\dagger}$, and Peter Stone$^{\dagger}$
\thanks{$^{\dagger}$Xuesu Xiao, Bo Liu, and Peter Stone are with Department of Computer Science, University of Texas at Austin, Austin, TX 78712 {\tt\scriptsize \{xiao, bliu, pstone\}@cs.utexas.edu}}
}

\begin{document}

\maketitle
\thispagestyle{empty}
\pagestyle{empty}

\begin{abstract}
Learning motion planners to move robot from one point to another within an obstacle-occupied space in a collision-free manner requires either an extensive amount of data or high-quality demonstrations. This requirement is caused by the fact that among the variety of maneuvers the robot can perform, it is difficult to find the single optimal plan without many trial-and-error or an expert who is already capable of doing so. However, given a plan performed in obstacle-free space, it is relatively easy to find an obstacle geometry, where this plan is optimal. We consider this ``dual'' problem of classical motion planning and name this process of finding appropriate obstacle geometry as \emph{hallucination}. In this work, we present two different approaches to hallucinate (1) the \emph{most constrained} and (2) a \emph{minimal} obstacle space where a given plan executed during an exploration phase in a completely safe obstacle-free environment remains optimal. We then train an end-to-end motion planner that can produce motions to move through realistic obstacles during deployment. Both methods are tested on a physical mobile robot in real-world cluttered environments. 
\end{abstract}

\section{INTRODUCTION}
\label{sec::intro}

While classical motion planners, such as Dynamic Window Approach (DWA) \cite{fox1997dynamic}, can reliably navigate the robot in cluttered spaces with properly tuned parameters, recent machine learning techniques have also been applied to the motion planning problem \cite{pfeiffer2017perception}. Those approaches either learns from a classical motion planner \cite{pfeiffer2017perception} or a human expert \cite{xiao2020appld}, or through an extensive amount of trial-and-error, such as Reinforcement Learning (RL) \cite{faust2018prm}. However, most learned motion planners still under-perform their classical counterparts. Despite the advantage of learning motion planners without hand-crafted rules \cite{fox1997dynamic} or in-situ adjustment \cite{xiao2017uav} in a data-driven fashion, the performance is still bottlenecked  by the requirement of good-quality training data \cite{liu2020lifelong}. 

Consider the ``dual" problem of motion planning: instead of finding the optimal motion plan for a specific obstacle configuration, either using hand-crafted rules or training data, we seek to find the obstacle configuration(s) where a specific motion plan is guaranteed to be optimal. We name this process \emph{hallucination}. Solving this problem gives us the freedom to allow random exploration in a completely safe obstacle-free space and collect an extensive amount of motion plans, whose optimally will be assured by a class of \emph{hallucination} techniques. In this work, we introduce two of those techniques: to hallucinate (1) the (unique) \emph{most constrained} and (2) a (not unique) \emph{minimal} obstacle configuration.

\section{GEOMETRIC HALLUCINATION}
\label{sec::approach}
Given a robot's configuration space (C-space) partitioned by unreachable (obstacle) and reachable (free) configurations, $C = C_{obst} \cup C_{free}$, we define the classical motion planning problem as to find a function $f(\cdot)$ that can be used to produce optimal plans $p=f(C_{obst}~|~c_c, c_g)$ that results in the robot moving from the robot's current configuration $c_c$ to a specified goal configuration $c_g$ without intersecting the interior of $C_{obst}$. Here, a plan $p \in \mathcal{P}$ is a sequence of low-level actions $\{u_i\}_{i=1}^{t}$, where $u_i\in\mathcal{U}$. This work introduces two methods to approach the ``dual" problem of finding optimal $f(\cdot)$. 

\subsection{Hallucinating Most Constrained Obstacle Space {\normalfont \cite{xiao2020toward}}}
\begin{figure}
\centering
\includegraphics[width=0.24\columnwidth]{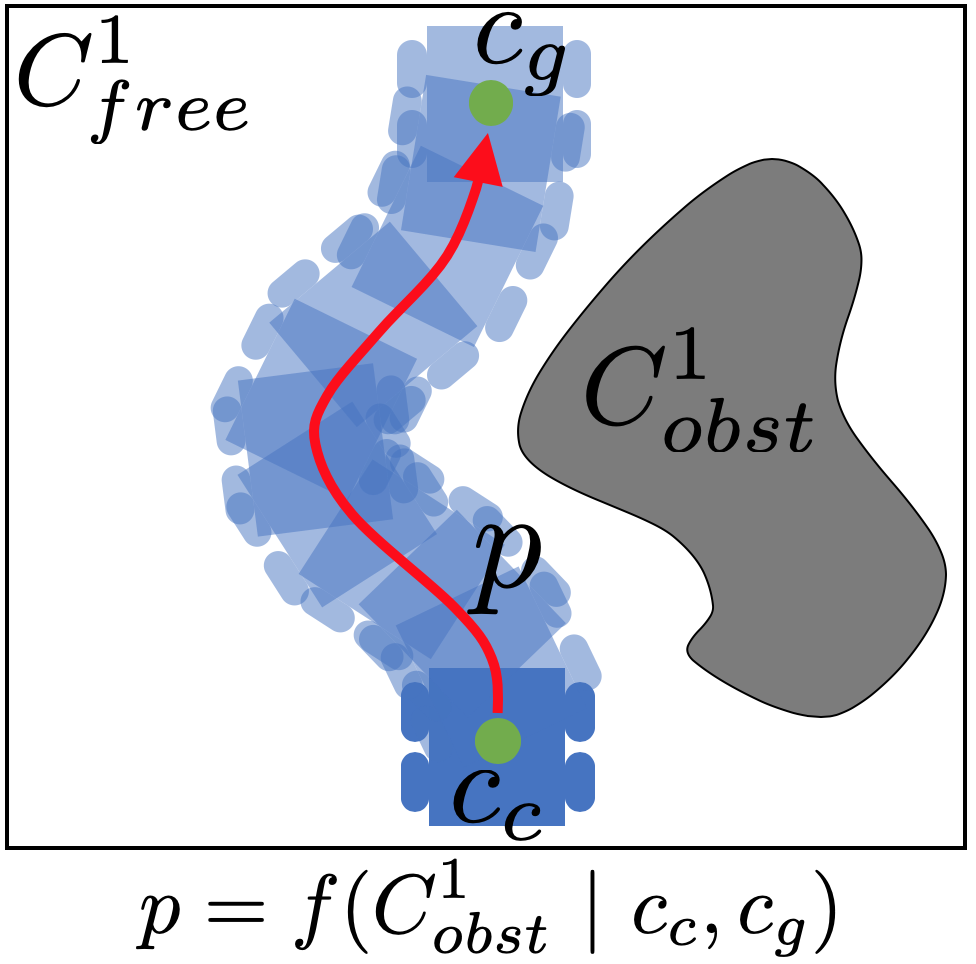}
\includegraphics[width=0.24\columnwidth]{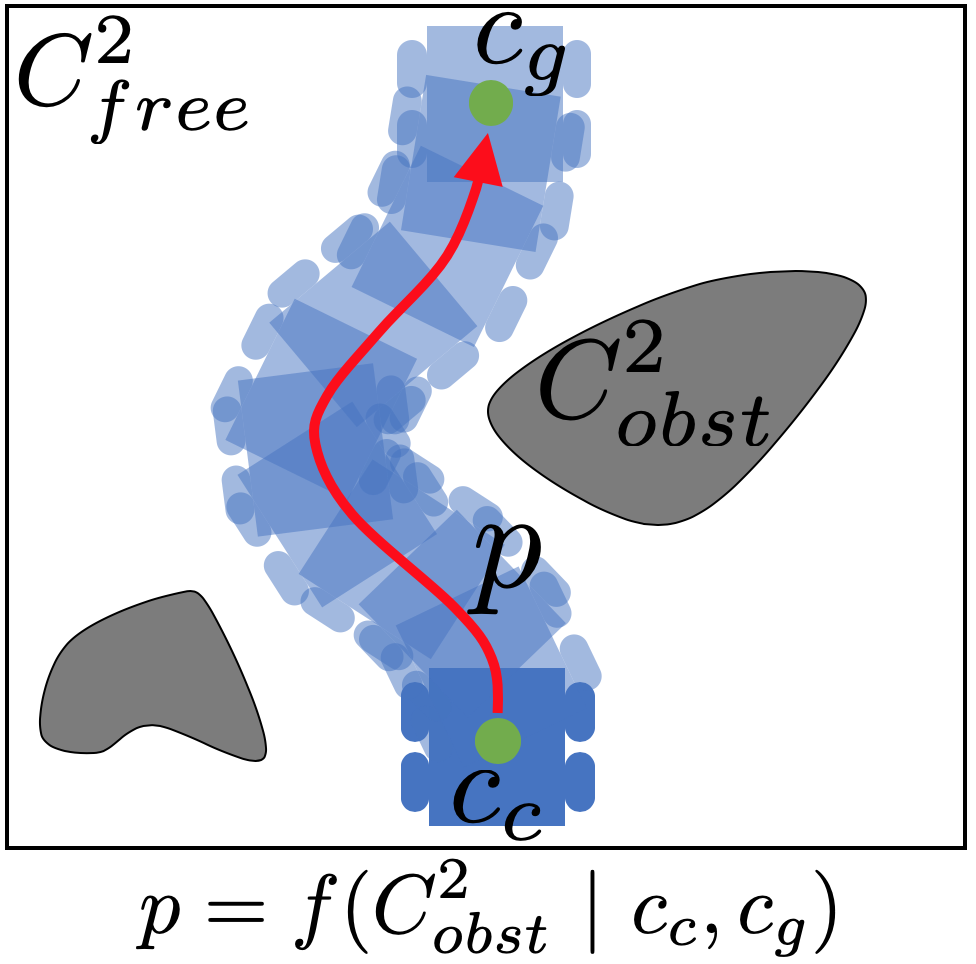}
\includegraphics[width=0.24\columnwidth]{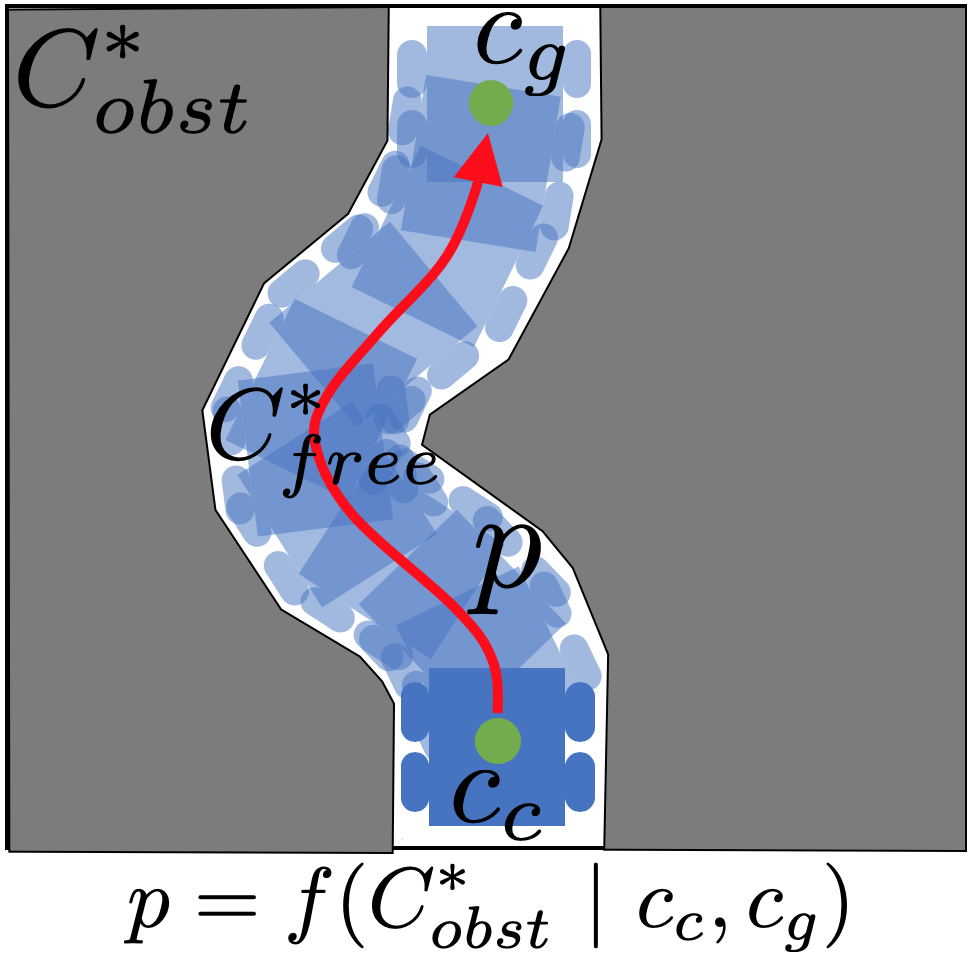}
\includegraphics[width=0.24\columnwidth]{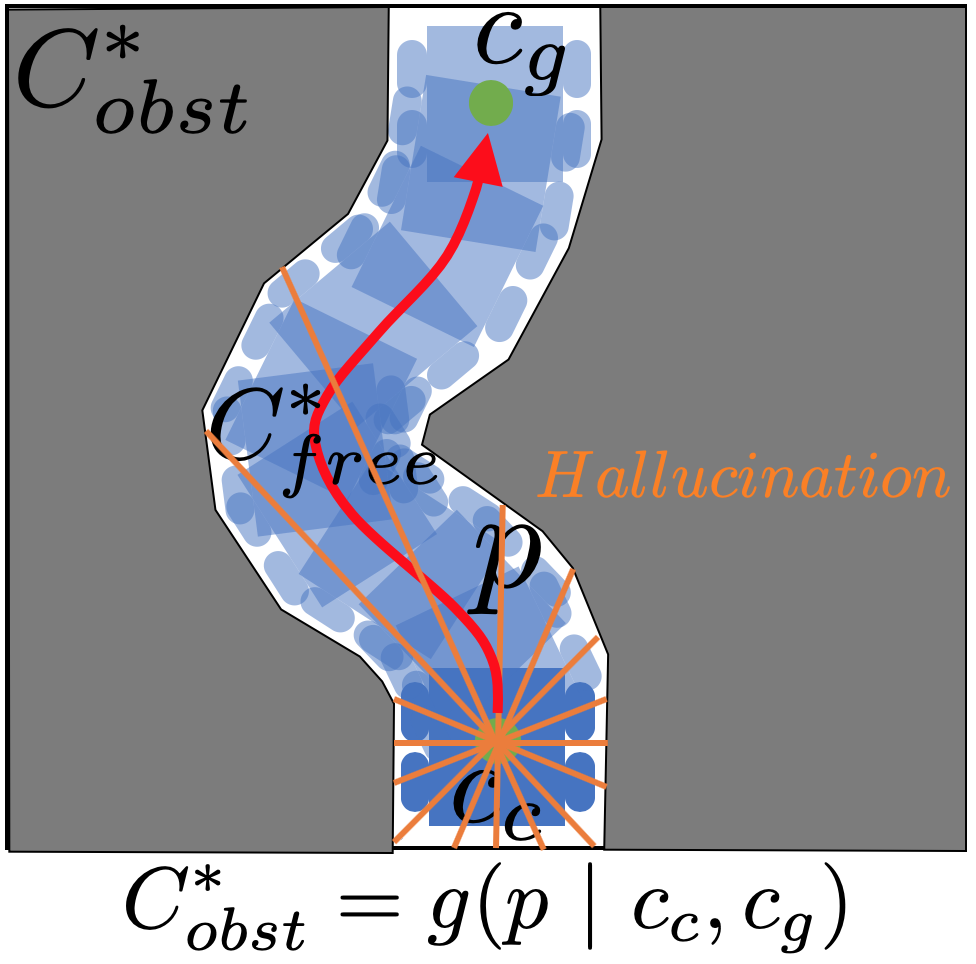}
\caption{$p=f(C_{obst}~|~c_c, c_g)$ and  $C_{obst}^*=g(p~|~c_c, c_g)$
}
\label{fig::halluci}
\vspace{-1pt}
\end{figure}

Since different $C_{obst}$ can lead to the same plan, the left inverse of $f$, $f^{-1}$, is not well defined (see Fig. \ref{fig::halluci} left). However, we can instead define a similar function $g(\cdot)$ such that $C_{obst}^* = g(p~|~c_c, c_g)$, where $C_{obst}^*$ denotes the C-space's {\em most constrained} unreachable set corresponding to $p$.\footnote{\scriptsize{Technically, $c_g$ can be uniquely determined by $p$ and $c_c$, but we include it as an input to $g(\cdot)$ for notational symmetry with $f(\cdot)$.}}  
Formally, given a plan $p$ and the set of all unreachable sets $\mathcal{C}_{obst}$, we say 
\begin{equation}
\begin{gathered}
C_{obst}^* = g(p~|~c_c, c_g) ~~\text{iff}~~\forall C_{obst}\in \mathcal{C}_{obst}, \\
f(C_{obst}~|~c_c, c_g) = p~~\Longrightarrow~~C_{obst} \subseteq C^*_{obst},
\end{gathered}
\end{equation}

We denote the corresponding reachable set of $C$ as $C_{free}^* = C \setminus C_{obst}^*  $. We call $g(\cdot)$ the \emph{most constrained} hallucination function and the output of $g(\cdot)$ a \emph{most constrained} hallucination. This hallucination can be projected onto the robot's sensors. For example, for a LiDAR sensor, we perform ray casting from the sensor to the boundary between $C_{obst}^*$ and $C_{free}^*$ in order to project the hallucination onto the range readings (Fig. \ref{fig::halluci} right). Given the hallucination $C_{obst}^*$ for $p$, the only viable (and therefore optimal) plan is $p=g^{-1}(C_{obst}^*~|~c_c, c_g)$. Note that $g(\cdot)$ is bijective and its inverse $g^{-1}(\cdot)$ is well defined. 
Leveraging machine learning, $g^{-1}(\cdot)$ is represented using a function approximator $g_{\theta}^{-1}(\cdot)$. Note that we aim to approximate $g_{\theta}^{-1}(\cdot)$ instead of the original $f(\cdot)$ due to the vastly different domain size: the \emph{most constrained} ($\mathcal{C}_{obst}^*$) vs. \emph{all} ($\mathcal{C}_{obst}$) unreachable sets. 

During deployment, we use a smoothed coarse global path from a global planner to generate runtime hallucination so $g_{\theta}^{-1}(\cdot)$ does not need to generalize to unseen scenarios. Other components, including a Turn in Place, Recovery Behavior, and Speed Modulation modules, are used in conjunction with $g_{\theta}^{-1}(\cdot)$ to address inevitable out-of-distribution scenarios and adapt to the real C-space. 


\subsection{Hallucinating Minimal Obstacle Space {\normalfont \cite{xiao2020agile}}}
\begin{figure}[t]
  \centering
  \includegraphics[height=90pt]{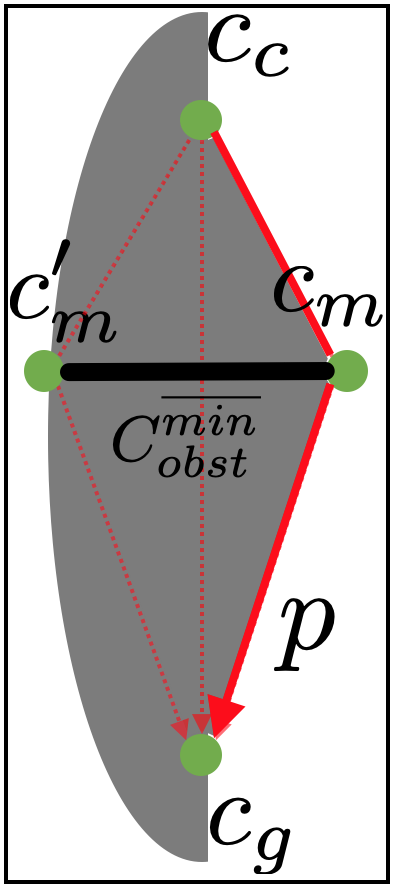}
  \includegraphics[height=90pt]{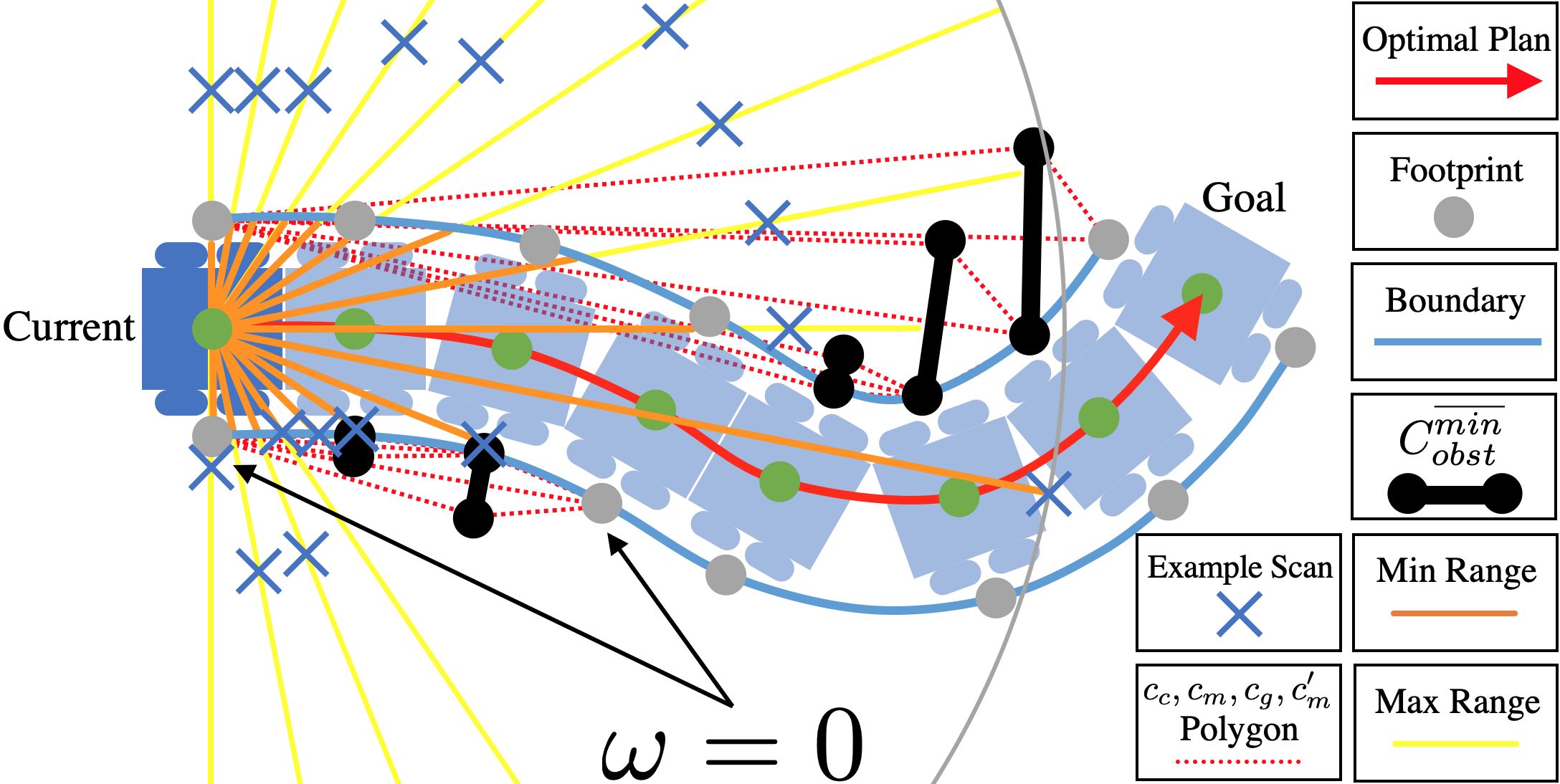}
  \caption{Left: $C_{obst}^{\overline{min}}$ is defined by three consecutive configurations with $c_c$, $c_m$, $c_g$, and symmetry point $c_m'$. Right: LiDAR reading is randomly sampled between min and max range. 
  }
  \label{fig::sober}
  \vspace{-15pt}
\end{figure}

Learning from hallucinated $C_{obst}^*$ can efficiently reduce input space, and therefore learning complexity, but requires runtime hallucination and other components during deployment. Hallucination of a \emph{minimal} obstacle space generates $C_{obst}^{min}$, which is a minimal set of obstacle configurations required to cause  the plan $p$ to be optimal, We then randomly samples augmentations to the minimal unreachable set. Formally, we define the set of $C_{obst}^{min}$ as: 
\begin{equation}
\begin{gathered}
\mathcal{C}_{obst}^{min} = \{C_{obst}^{min} \mid \forall c \in C_{obst}^{min}, \\
f(C_{obst}^{min}\setminus\{c\}~|~c_c, c_g) \neq f(C_{obst}^{min}~|~c_c,c_g)\}
\end{gathered}
\end{equation}
We use a special $C_{obst}^{\overline{min}}$ to approximate any $C_{obst}^{min} \in \mathcal{C}_{obst}^{min}$ (Fig. \ref{fig::sober} left). This approximation is sufficient when the robot trajectory is composed of a dense sequence of configurations and $C_{obst}^{min}$ is instantiated on discrete LiDAR beams, which will be shown empirically. As shown in Fig. \ref{fig::sober} right, the max range of a LiDAR beam is determined by $C_{obst}^{\overline{min}}$ (if the beam intersects $C_{obst}^{\overline{min}}$) or the sensor's physical limit (if not), while the min range for each beam is determined by the boundary of the robot path. 
A random range is sampled between the min and max values, considering possible continuity among neighboring beams and being offset for uncertainty/safety induced by the optimal plan $p$. Therefore, many $C_{obst}$ can be augmented based on $C_{obst}^{\overline{min}}$. We then train a parameterized policy $f_{\theta}(\cdot)$ to approximate classical motion planner $f(\cdot)$. 

The advantage of augmenting $C_{obst}^{\overline{min}}$  and generating many $C_{obst}$ is, during deployment, no runtime hallucination with the help of a global path and other extra components are required. The learned $f_{\theta}(\cdot)$ can plan in response to the real perception and adapt to the actual scenarios on its own. 

\subsection[]{Physical Experiments\footnote{\scriptsize{Videos: 
\url{https://www.youtube.com/watch?v=T72Z6rz9ges&t=1s} and \url{https://www.youtube.com/watch?v=xtLaSF0kiB0&t=49s}. }}}

Two datasets are collected by two random exploration policies in an obstacle-free space in simulation: one with mostly constant 0.4m/s linear velocity ($v\approx 0.4$m/s) and varying angular velocity ($\omega \in [-1.57, 1.57]$rad/s), the other with varying $v \in [0, 1.0]$m/s and $\omega \in [-1.57, 1.57]$rad/s. If trained on the first dataset, the speed of the planner output is modulated by a Model Predictive Control based collision probability checker, achieving a max $v=0.6m/s$. Four neural network based planners are trained using the two datasets and two hallucination techniques. Simulated \cite{perille2020benchmarking} and physical experiments are performed. While the \emph{minimal} hallucination works well on both datasets and outperforms all other variants, and even a classical motion planner \cite{fox1997dynamic}, the \emph{most constrained} hallucination only performs well on the 0.4m/s dataset. This is because learning from varying speed while hallucinating only the most constrained space causes ambiguity for the learner.

\section{CONCLUSIONS}
\label{sec::conclusions}
We present a class of two geometric hallucination techniques that approach the classical motion planning problem from the opposite direction. Instead of seeking an optimal motion plan for an obstacle configuration, we find the obstacle configuration(s), where a motion plan is optimal. The first approach hallucinates the \emph{most constrained} C-space, where the plan is the only feasible, and therefore optimal, plan. It largely reduces the learning complexity, since the learned motion planner $g^{-1}_\theta(\cdot)$ only plans in the most constrained C-spaces, instead of any C-spaces. However, the downside of this approach is during deployment, runtime hallucination along with other extra components are necessary. The second approach finds a \emph{minimal} obstacle set to make a plan optimal, augments this minimal set to generate a large body of training data, and therefore does not require any extra components during deployment.

\bibliographystyle{IEEEtran}
\bibliography{IEEEabrv,references}

\end{document}